\documentclass[11pt,a4paper]{article}
\usepackage[hyperref]{acl2020}
\usepackage{times}
\usepackage{latexsym}
\usepackage{amsfonts}
\usepackage{bm}
\usepackage{comment}
\usepackage{booktabs}
\usepackage{multirow}
\usepackage{colortbl}
\usepackage{graphicx}
\usepackage[normalem]{ulem}
\useunder{\uline}{\ul}{}\usepackage{relsize}

\usepackage{microtype}

\aclfinalcopy % Uncomment this line for the final submission
\let\vec\mathbf

\title{Accelerating Natural Language Understanding\\in Task-Oriented Dialog}

\author{Ojas Ahuja \and Shrey Desai \\
  Department of Computer Science \\
  The University of Texas at Austin \\
  \texttt{\{ojas, shreydesai\}@utexas.edu}}

\date{}

\begin{document}
\maketitle
\begin{abstract}
Task-oriented dialog models typically leverage complex neural architectures and large-scale, pre-trained Transformers to achieve state-of-the-art performance on popular natural language understanding benchmarks. However, these models frequently have in excess of tens of millions of parameters, making them impossible to deploy on-device where resource-efficiency is a major concern. In this work, we show that a simple convolutional model compressed with structured pruning achieves largely comparable results to BERT \cite{Devlin2019BERTPO} on ATIS and Snips, with under 100K parameters. Moreover, we perform acceleration experiments on CPUs, where we observe our multi-task model predicts intents and slots nearly 63$\times$ faster than even DistilBERT \citep{Sanh2019DistilBERTAD}. 
\end{abstract}

\section{Introduction}

The advent of smart devices like Amazon Alexa, Facebook Portal, and Google Assistant has increased the necessity of resource-efficient task-oriented systems \citep{Coucke2018SnipsVP,zhang2020recent,desai2020lightweight}. These systems chiefly perform two natural language understanding tasks, intent detection and slot filling, where the goals are to understand what the user is trying to accomplish and the metadata associated with the request, respectively \citep{gupta-etal-2018-semantic}. However, there remains a disconnect between state-of-the-art task-oriented systems and their deployment in real-world applications. Recent top performing systems have largely saturated performance on ATIS \citep{hemphill1990atis} and Snips \citep{Coucke2018SnipsVP} by leveraging complex neural architectures and large-scale, pre-trained language models \citep{Devlin2019BERTPO}, but their usability in on-device settings remains suspect \citep{Qin2019ASF,Cheng2017ASO}. Mobile phones, for example, have sharp hardware constraints and limited memory capacities, implying systems must optimize for both accuracy and resource-efficiency as possible to be able to run in these types of environments \citep{Lin2010TuningMA,McIntosh2018WhatCA}.

In this work, we present a vastly simplified, single-layer convolutional model \citep{Kim2014ConvolutionalNN, Bai2018AnEE} that is highly compressible but nonetheless achieves competitive results on task-oriented natural language understanding benchmarks. In order to compress the model, we use structured magnitude-based pruning \citep{Anwar2017StructuredPO,li2017pruning}, a two-step approach where (1) entire convolutional filters are deleted according to their $\ell_2$ norms; and (2) remaining portions of the underlying weight matrix are spliced together. The successive reduction in the number of convolutional output connections permits downstream weight matrices to reduce their number of input connections as well, collectively resulting in a smaller model. Structured pruning and re-training steps are then interleaved to ensure the model is able to reconstruct lost filters that may contribute valuable information. During test-time, however, we use the pruned model as-is without further fine-tuning.

Our simple convolutional model with structured pruning obtains strong results despite having less than 100K parameters. On ATIS, our multi-task model achieves 95\% intent accuracy and 94\% slot F1, only about 2\% lower than BERT \cite{Devlin2019BERTPO}. Structured pruning also admits significantly faster inference: on CPUs, we show our model is 63$\times$ faster than DistilBERT. Unlike compression methods based on unstructured pruning \cite{frankle2019lottery}, our model enjoys speedups \textit{without} having to rely on a sparse tensor library at test-time \cite{Han2015DeepCC}, thus we demonstrate the potential for usage in resource-constrained, on-device settings. Our code is publicly available at \url{https://github.com/oja/pruned-nlu}.

\section{Related Work}

\paragraph{Task-Oriented Dialog.} Dialog systems perform a range of tasks, including language understanding, dialog state tracking, content planning, and text generation \citep{Bobrow1977GUSAF,henderson-machine-2015,yu-etal-2016-strategy,Yan2017BuildingTD,gao-etal-2018-neural-approaches}. For smart devices, specifically, intent detection and slot filling form the backbone of natural language understanding (NLU) modules, which can either be used in single-turn or multi-turn conversations \citep{Coucke2018SnipsVP,rastogi2019towards}. We contribute a single-turn, multi-task NLU system especially tailored for on-device settings, as demonstrated through acceleration experiments.

\paragraph{Model Compression.} In natural language processing, numerous works have used compression techniques like quantization \citep{Wrbel2018ConvolutionalNN,zafrir2019q8bert}, distillation \citep{Sanh2019DistilBERTAD,tang2019distilling,jiao2020tinybert}, pruning \citep{yoon2018,gordon2020compressing}, and smaller representations \citep{raviselfgoverning,kozarevafast,desai2020lightweight}. Concurrently, \citet{desai2020lightweight} develop lightweight convolutional representations for on-device task-oriented systems, related to our goals, but they do not compare with other compression methods and solely evaluate on a proprietary dataset. In contrast, we compare the efficacy of structured pruning against strong baselines---including BERT \citep{Devlin2019BERTPO}---on the open-source ATIS and Snips datasets.

\section{Convolutional Model}

\paragraph{Convolutions for On-Device Modeling.} State-of-the-art task-oriented models are largely based on recurrent neural networks (RNNs) \citep{Wang2018ABB} or Transformers \citep{Qin2019ASF}. However, these models are often impractical to deploy in low-resource settings. Recurrent models must sequentially unroll sequences during inference, and self-attention mechanisms in Transformers process sequences with quadratic complexity \citep{Vaswani2017AttentionIA}. High-performing, pre-trained Transformers, in particular, also have upwards of tens of millions of parameters, even when distilled \citep{tang2019distilling,Sanh2019DistilBERTAD}.

Convolutional neural networks (CNNs), in contrast, are highly parallelizable and can be significantly compressed with structured pruning \citep{li2017pruning}, while still achieving competitive performance on a variety of NLP tasks \citep{Kim2014ConvolutionalNN,Gehring2017ConvolutionalST}. Furthermore, the core convolution operation has enjoyed speedups with dedicated digital signal processors (DSPs) and field programmable gate arrays (FPGAs) \citep{ahmad2020FPGA}. Model compatibility with on-device hardware is one of the most important considerations for practitioners as, even if a model works well on high throughput GPUs, its components may saturate valuable resources like memory and power \citep{Lin2010TuningMA}.

\paragraph{Model Description.} Model inputs are encoded as a sequence of integers $\mathbf{w}=(w_1, \cdots, w_n)$ and right-padded up to a maximum sequence length. The embedding layer replaces each token $w_i$ with a corresponding $d$-dimensional vector $\vec{e}_i \in \mathbb{R}^{d}$ sourced from pre-trained GloVe embeddings \citep{pennington2014glove}. A feature map $\vec{c} \in \mathbb{R}^{n-h+1}$ is then calculated by applying a convolutional filter of height $h$ over the embedded input sequence. We apply max-over-time pooling $\hat{c}=\mathrm{max}(\mathbf{c})$ \citep{Collobert2011NaturalLP} to simultaneously reduce the dimensionality and extract the most salient features. The pooled features are then concatenated and fed through a linear layer with dropout \citep{Srivastava2014DropoutAS}. The objective is to maximize the log likelihood of intents, slots, or both (under a multi-task setup), and is optimized with Adam \citep{Kingma2015AdamAM}.

To ensure broad applicability, our model emphasizes simplicity, and therefore minimizes the number of extraneous architectural decisions: there is only a single convolutional block, no residual connections, and no normalization layers.

\paragraph{Temporal Padding.} The model described above is capable of predicting an intent that encompasses the entire input sequence, but cannot be used for sequence labeling tasks, namely slot filling. To create a one-to-one correspondence between input tokens and output slots, \citet{Bai2018AnEE} left-pad the input sequence by $k - 1$, where $k$ is the kernel size. We modify this by loosening the causality constraint and instead padding each side by $\frac{k-1}{2}$. Visually, this results in a ``centered'' convolution that inculcates bidirectional context when computing a feature map. Note that this padding is unnecessary for intent detection, therefore we skip it when training a single-task intent model.

\paragraph{Multi-Task Training.} Intent detection and slot filling can either be disjointly learned with dedicated single-task models or jointly learned with a unified multi-task model \citep{Liu2016AttentionBasedRN}. In the latter model, we introduce task-specific heads on top of the common representation layer and simultaneously optimize both objectives:
$$\mathcal{L}_\mathrm{joint} = \alpha \mathcal{L}_\mathrm{intent} + (1-\alpha)\mathcal{L}_\mathrm{slot}$$
for $\alpha$ where $0 \leq \alpha \leq 1$. Empirically, we observe that weighting $\mathcal{L}_\mathrm{slot}$ more than $\mathcal{L}_\mathrm{intent}$ results in higher performance ($\alpha \approx 0.2$). Our hypothesis is that, because of the comparative difficulty of the slot filling task, the model is required to learn a more robust representation of each utterance, which is nonetheless useful for intent detection.

\section{Structured Pruning}

\paragraph{Structured vs. Unstructured Pruning.} Pruning is one compression technique that removes weights from an over-parameterized model \citep{lecun1990optimal}, often relying on a heuristic function that ranks weights (or groups of weights) by their importance. Methods for pruning are broadly categorized as unstructured and structured: unstructured pruning allows weights to be removed haphazardly without geometric constraints, but structured pruning induces well-defined sparsity patterns, for example, dropping entire filters in a convolutional layer according to their norm \citep{Molchanov2016PruningCN, li2017pruning, Anwar2017StructuredPO}. Critically, \textbf{the model's true size is not diminished with unstructured pruning}, as without a sparse tensor library, weight matrices with scattered zero \textit{elements} must still be stored \cite{Han2015DeepCC}. In contrast, structurally pruned models do not rely on such libraries at test-time since non-zero \textit{units} can simply be spliced together.

\begin{figure}[t]
    \centering
    \includegraphics[width=\hsize]{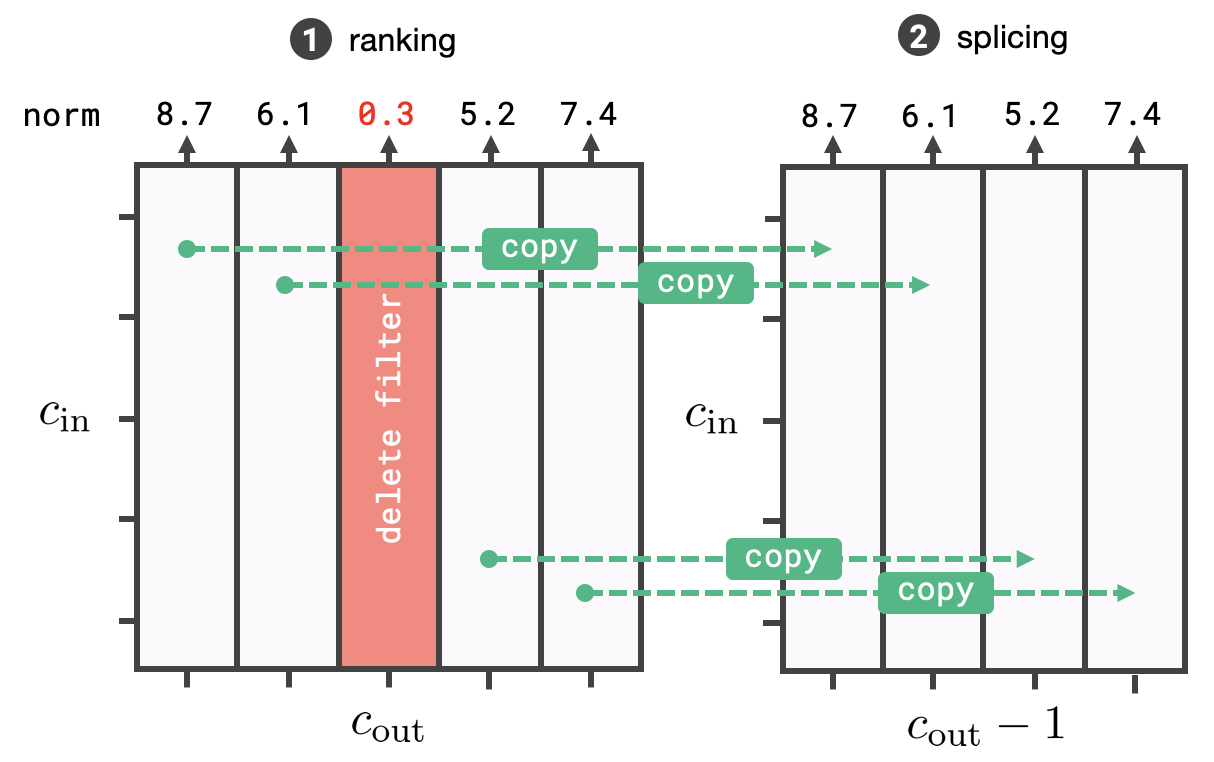}
    \caption{Structured pruning of convolutional models by (1) ranking filters by their $\ell_2$ norm, then (2) splicing out the lowest norm filter, resulting in a successively smaller weight matrix. Because each filter convolves input filters $c_\mathrm{in}$ into one output filter $c_\mathrm{out}$, removing a single filter results in $c_\mathrm{out}-1$ output channels.}
    \label{fig:schematic}
\end{figure}

\paragraph{Pruning Methodology.} The structured pruning process is depicted in Figure~\ref{fig:schematic}. In each pruning iteration, we rank each filter by its $\ell_2$ norm, greedily remove filters with the smallest magnitudes, and splice together non-zero filters in the underlying weight matrix. The deletion of a single filter results in one less output channel, implying we can also remove the corresponding input channel of the subsequent linear layer with a similar splicing operation. Repetitions of this process result in an objectively smaller model because of reductions in the convolutional and linear layer weight matrices. Furthermore, this process does not lead to irregular sparsity patterns, resulting in a general speedup on all hardware platforms.

The heuristic function for ranking filters and whether to re-train the model after a pruning step are important hyperparameters. We experimented with both $\ell_1$ and $\ell_2$ norms for selecting filters, and found that $\ell_2$ slightly outperforms $\ell_1$. More complicated heuristic functions, such as deriving filter importance according to gradient saliency \citep{Persand2020CompositionOS}, can also be dropped into our pipeline without modification. 

\paragraph{One-Shot vs. Iterative Pruning.} Furthermore, when deciding to re-train the model, we experiment with one-shot and iterative pruning \citep{frankle2019lottery}. One-shot pruning involves repeatedly deleting filters until reaching a desired sparsity level without re-training, whereas iterative pruning interleaves pruning and re-training, such that the model is re-trained to convergence after each pruning step. These re-training steps increase overall training time, but implicitly help the model ``reconstruct'' deleted filter(s), resulting in significantly better performance. During test-time, however, the pruned model uses significantly fewer resources, as we demonstrate in our acceleration experiments.

\section{Tasks and Datasets}

We build convolutional models for intent detection and slot filling, two popular natural language understanding tasks in the task-oriented dialog stack. Intent detection is a multi-class classification problem, whereas slot filling is a sequence labeling problem. Formally, given utterance tokens $\mathbf{w}=(w_1, \cdots, w_n)$, models induce a joint distribution $P(y^*_{\mathrm{intent}}, \mathbf{y}^*_\mathrm{slot} | \mathbf{w})$ over an intent label $y^*_\mathrm{intent}$ and slot labels $\mathbf{y}^*_\mathrm{slot}=(\mathbf{y}^{*(1)}_\mathrm{slot}, \cdots, \mathbf{y}^{*(n)}_\mathrm{slot})$. These models are typically multi-task: intent and slots predictions are derived with task-specific heads but share a common representation \cite{Liu2016AttentionBasedRN}. However, since the intent and slots of an utterance are independent, we can also learn single-task models, where an intent model optimizes $P(y^*_{\mathrm{intent}} | \mathbf{w})$ and a slot model optimizes $P(\mathbf{y}^*_\mathrm{slot} | \mathbf{w})$. We experiment with both approaches, although our ultimate compressed model is multi-tasked as aligned with on-device use cases.

Following previous work, we evaluate on ATIS \citep{hemphill1990atis} and Snips \citep{Coucke2018SnipsVP}, both of which are single-turn dialog benchmarks with intent detection and slot filling. ATIS has 4,478/500/893 train/validation/test samples, respectively, with 21 intents and 120 slots. Snips has 13,084/700/700 samples with 7 intents and 72 slots. Our setup follows the same preparation as \citet{zhang2019joint}.

\begin{table}[t]
\centering
\small
\setlength{\tabcolsep}{4pt}
\begin{tabular}{lrrrrrr}
\toprule
\multirow{2}{*}{Models}
& \multicolumn{2}{c}{ATIS} & \multicolumn{2}{c}{Snips} \\
\cmidrule(lr){2-3} \cmidrule(lr){4-5}
& Intent & Slot & Intent & Slot \\
\midrule
\multicolumn{5}{l}{\textbf{Baselines}} \\
\midrule
Slot-Gated RNN & 94.10 & 95.20 & 97.00 & 88.80 \\
Stack Propagation & 96.90 & 95.90 & 98.00 & 94.20 \\
DistilBERT (66M) & 96.98 & 95.44 & 97.94 & 94.59 \\
BERT (110M) & 97.16 & 96.02 & 98.26 & 95.05 \\
\midrule
\multicolumn{5}{l}{\textbf{Method: No Compression}} \\
\midrule
Single-Task & 94.94 & 94.01 & 96.54 & 85.06 \\
Multi-Task (195K/174K) & 94.98 & 94.30 & 96.97 & 84.38 \\
\midrule
\multicolumn{5}{l}{\textbf{Method: Structured Pruning}} \\
\midrule
Single-Task & 95.45 & 94.61 & 96.94 & 85.11 \\
Multi-Task (97K/87K) & 95.39 & 94.42 & 97.17 & 83.81 \\
\bottomrule
\end{tabular}
\caption{Intent accuracy and slot F1 of baseline models \citep{Goo2018SlotGatedMF,Qin2019ASF,Sanh2019DistilBERTAD,Devlin2019BERTPO} and our systems on ATIS and Snips. We experiment with single-task and multi-task models. Number of model parameters are shown in parentheses where applicable; multi-task models use the format (ATIS/Snips).}
\label{tab:performance}
\end{table}

\section{Experiments and Results}

We evaluate the performance, compression, and acceleration of our structured pruning approach against several baselines. Note that we do not employ post-hoc compression methods like quantization \citep{Guo2018ASO}, as they are orthogonal to our core method, and can be utilized at no additional cost to further improve performance on-device. 

\subsection{Benchmark Results}
    
We experiment with both single-task and multi-task models, with and without structured pruning, on ATIS and Snips. The results are displayed in Table \ref{tab:performance}. Our multi-task model with structured pruning, even with over a 50\% reduction in parameters, performs on par with our \textsc{no compression} baselines. On ATIS, our model is comparable to \textsc{slot-gated rnn} \citep{Goo2018SlotGatedMF} and is only about 2\% worse in accuracy/F1 than BERT. However, we note that our model's slot F1 severely drops off on Snips, possibly because it is a much larger dataset spanning a myriad of domains. Whether our pre-trained embeddings have sufficient explanatory power to scale past common utterances is an open question. 

Furthermore, to approximate what information is lost after compression, we analyze which samples' predictions flip from correct to incorrect after structured pruning. We observe that sparser models tend to prefer larger classes; for example, in slot filing, tags are often mislabeled as ``outside'' in IOB labeling \cite{sang-2000-bio}. This demonstrates a trade-off between preserving non-salient features that work on average for all classes or salient features that accurately discriminate between the most prominent classes. Our model falls on the right end of this spectrum, in that it greedily de-prioritizes representations for inputs that do not contribute as much to aggregate dataset log likelihood.

\begin{table}[t]
\centering
\small
\setlength{\tabcolsep}{4pt}
\begin{tabular}{rrrrrrrr}
\toprule
\multirow{2}{*}{Params} & \multirow{2}{*}{CR (\%)} & \multicolumn{2}{c}{Pruning} & \multicolumn{2}{c}{Distillation} \\
\cmidrule(lr){3-4} \cmidrule(lr){5-6}
& & Intent & Slot & Intent & Slot \\
\midrule
195K & 0\% & \cellcolor{red!5.02}{94.98} & \cellcolor{red!5.70}{94.30} & \cellcolor{red!6.16}{93.84} & \cellcolor{red!5.88}{94.12} \\
156K & 20\% & \cellcolor{red!4.61}{95.39} & \cellcolor{red!5.81}{94.19} & \cellcolor{red!5.15}{94.85} & \cellcolor{red!5.78}{94.22} \\
117K & 40\% & \cellcolor{red!4.97}{95.03} & \cellcolor{red!5.86}{94.14} & \cellcolor{red!5.49}{94.51} & \cellcolor{red!5.87}{94.13} \\
78K & 60\% & \cellcolor{red!4.90}{95.10} & \cellcolor{red!5.88}{94.12} & \cellcolor{red!7.73}{92.27} & \cellcolor{red!5.68}{94.32} \\
39K & 80\% & \cellcolor{red!5.60}{94.40} & \cellcolor{red!6.09}{93.91} & \cellcolor{red!9.52}{90.48} & \cellcolor{red!5.95}{94.05} \\
19K & 90\% & \cellcolor{red!7.77}{92.23} & \cellcolor{red!6.80}{93.20} & \cellcolor{red!21.72}{78.28} & \cellcolor{red!7.54}{92.46} \\
9K & 95\% & \cellcolor{red!11.65}{88.35} & \cellcolor{red!7.81}{92.19} & \cellcolor{red!29.11}{70.89} & \cellcolor{red!10.46}{89.54} \\
2K & 99\% & \cellcolor{red!20.51}{79.49} & \cellcolor{red!12.83}{87.17} & \cellcolor{red!29.11}{70.89} & \cellcolor{red!35.25}{64.75} \\
\bottomrule
\end{tabular}
\caption{ATIS performance of multi-task models compressed with structured pruning (ours) and knowledge distillation \cite{Hinton2015DistillingTK} as the compression rate (CR; \%) increases. We report intent accuracy and slot F1. Darker shades of \colorbox{red!50}{red} indicate higher absolute performance drops with respect to 100\%.}
\label{tab:comparison}
\end{table}

\iffalse
\begin{table}[t]
\centering
\small
\setlength{\tabcolsep}{4pt}
\begin{tabular}{lrrrrrr}
\toprule
\multirow{2}{*}{Params Left (\%)}
& \multicolumn{2}{c}{Pruning} & \multicolumn{2}{c}{Distillation} \\
\cmidrule(lr){2-3} \cmidrule(lr){4-5}
& Intent & Slot & Intent & Slot \\
\midrule
195K (0\%) & 94.98 & 94.30 & 93.84 & 94.12 \\
156K (20\%) & 95.39 & 94.19 & 94.85 & 94.22 \\
117K (40\%) & 95.03 & 94.14 & 94.51 & 94.13 \\
78K (60\%) & 95.10 & 94.12 & 92.27 & 94.32 \\
39K (80\%) & 94.40 & 93.91 & 90.48 & 94.05 \\
19K (90\%) & 92.23 & 93.20 & 78.28 & 92.46 \\
9K (95\%) & 88.35 & 92.19 & 70.89 & 89.54 \\
2K (99\%) & 79.49 & 87.17 & 70.89 & 64.75 \\
\bottomrule
\end{tabular}
\caption{Performance-compression tradeoff curves on ATIS and Snips, comparing multi-task models compressed with structured pruning (ours) and knowledge distillation \cite{Hinton2015DistillingTK}.}
\label{tab:comparison}
\end{table}
\fi

\subsection{Comparison with Distillation}

\begin{figure}
    \centering
    \includegraphics[width=\hsize]{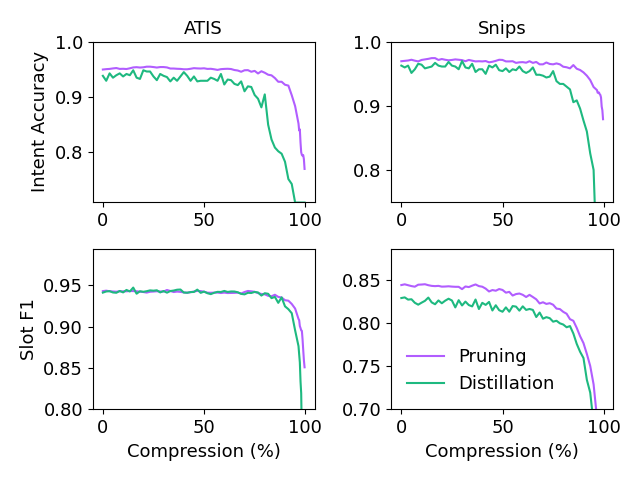}
    \caption{Performance-compression tradeoff curves on ATIS and Snips, comparing multi-task models compressed with structured pruning (ours) and knowledge distillation \cite{Hinton2015DistillingTK}. Pruning curves denote the mean of five compression runs with random restarts. Note that the $y$-axis ticks are \textit{not} uniform across graphs.}
    \label{fig:pruning_vs_distillation}
\end{figure}

In addition, we compare structured pruning with knowledge distillation, a popular compression technique where a small, student model learns from a large, teacher model by minimizing the KL divergence between their output distributions \citep{Hinton2015DistillingTK}. Using a multi-task model on ATIS, we compress it with structured pruning and distillation, then examine its performance at varying levels of compression. The results are shown in Table \ref{tab:comparison}. Distillation achieves similar results as structured pruning with 0-50\% sparsity, but its performance largely drops off after 80\%. Surprisingly, even with extreme compression (99\%), structured pruning is about 10\% and 20\% better on intents and slots, respectively.

Our results show that, in this setting, the iterative refinement of a sparse topology admits an easier optimization problem; learning a smaller model directly is not advantageous, even when it is supervised with a larger model. Furthermore, the iterative nature of structured pruning means it is possible to select a model that optimizes a particular performance-compression trade off along a sparsity curve, as shown in Figure \ref{fig:pruning_vs_distillation}. To do the same with distillation requires re-training for a target compression level each time, which is intractable with a large set of hyperparameters.

\begin{table}[t]
\centering
\small
\setlength{\tabcolsep}{4pt}
\begin{tabular}{lrrrrrr}
\toprule
\multirow{2}{*}{System}
& \multicolumn{2}{c}{ATIS} & \multicolumn{2}{c}{Snips} \\
\cmidrule(lr){2-3} \cmidrule(lr){4-5}
& CPU $\downarrow$  & GPU $\downarrow$ & CPU $\downarrow$  & GPU $\downarrow$  \\
\midrule
DistilBERT & 22.15 ms & 1.87 ms & 21.81 ms & 1.76 ms \\
BERT & 43.19 ms & 2.80 ms & 43.04 ms & 2.72 ms \\
\midrule
Pruning & \textbf{0.35} ms & \textbf{0.37} ms & \textbf{0.33} ms & \textbf{0.36} ms \\
Distillation & 0.40 ms & 0.37 ms & 0.38 ms & 0.37 ms \\
\bottomrule
\end{tabular}
\caption{Average CPU and GPU inference time (in milliseconds) of baselines \citep{Sanh2019DistilBERTAD,Devlin2019BERTPO} and our multi-task models on ATIS and Snips.}
\label{tab:acceleration}
\end{table}

\subsection{Acceleration Experiments}

Lastly, to understand how our multi-task model with structured pruning performs without significant computational resources, we benchmark its test-time performance on a CPU and GPU. Specifically, we measure several models' inference times on ATIS and Snips (normalized by the total number of test samples) using an Intel Xeon E3-1270 v3 CPU and NVIDIA GTX 1080-TI GPU. Results are shown in Table \ref{tab:acceleration}. Empirically, we see that our pruned model results in significant speedups without a GPU compared to both a distilled model and BERT. DistilBERT, which is a strong approximation of BERT, is still 63$\times$ slower than our model. We expect that latency disparities on weaker CPUs will be even more extreme, therefore selecting a model that maximizes both task performance and resource-efficiency will be an important consideration for practitioners.

\section{Conclusion}

In this work, we show that structurally pruned convolutional models achieve competitive performance on intent detection and slot filling at only a fraction of the size of state-of-the-art models. Our method outperforms popular compression methods, such as knowledge distillation, and results in large CPU speedups compared to BERT and DistilBERT.

\section*{Acknowledgments}

Thanks to our anonymous reviewers for their helpful comments and feedback.

\bibliography{acl2020}

\begin{thebibliography}{48}
\expandafter\ifx\csname natexlab\endcsname\relax\def\natexlab#1{#1}\fi

\bibitem[{Ahmad and Pasha(2020)}]{ahmad2020FPGA}
Afzal Ahmad and Muhammad~Adeel Pasha. 2020.
\newblock {FFConv: An FPGA-Based Accelerator for Fast Convolution Layers in
  Convolutional Neural Networks}.
\newblock \emph{ACM Transactions on Embedded Computing Systems}.

\bibitem[{Anwar et~al.(2017)Anwar, Hwang, and Sung}]{Anwar2017StructuredPO}
Sajid Anwar, Kyuyeon Hwang, and Wonyong Sung. 2017.
\newblock {Structured Pruning of Deep Convolutional Neural Networks}.
\newblock \emph{ACM Journal on Emerging Technologies in Computing Systems
  (JETC)}.

\bibitem[{Bai et~al.(2018)Bai, Kolter, and Koltun}]{Bai2018AnEE}
Shaojie Bai, J.~Zico Kolter, and Vladlen Koltun. 2018.
\newblock {An Empirical Evaluation of Generic Convolutional and Recurrent
  Networks for Sequence Modeling}.
\newblock \emph{arXiv preprint arXiv:1803.01271}.

\bibitem[{Bobrow et~al.(1977)Bobrow, Kaplan, Kay, Norman, Thompson, and
  Winograd}]{Bobrow1977GUSAF}
Daniel~G. Bobrow, Ronald~M. Kaplan, Martin Kay, Donald~A. Norman, Henry~S.
  Thompson, and Terry Winograd. 1977.
\newblock {GUS, A Frame-Driven Dialog System}.
\newblock \emph{Artificial Intelligence}.

\bibitem[{Cheng et~al.(2017)Cheng, Wang, Zhou, and Zhang}]{Cheng2017ASO}
Yu~Cheng, Duo Wang, Pan Zhou, and Tao Zhang. 2017.
\newblock {A Survey of Model Compression and Acceleration for Deep Neural
  Networks}.
\newblock \emph{arXiv preprint arXiv:1710.09282}.

\bibitem[{Collobert et~al.(2011)Collobert, Weston, Bottou, Karlen, Kavukcuoglu,
  and Kuksa}]{Collobert2011NaturalLP}
Ronan Collobert, Jason Weston, L{\'e}on Bottou, Michael Karlen, Koray
  Kavukcuoglu, and Pavel~P. Kuksa. 2011.
\newblock {Natural Language Processing (Almost) from Scratch}.
\newblock \emph{Journal of Machine Learning Research (JMLR)}.

\bibitem[{Coucke et~al.(2018)Coucke, Saade, Ball, Bluche, Caulier, Leroy,
  Doumouro, Gisselbrecht, Caltagirone, Lavril, Primet, and
  Dureau}]{Coucke2018SnipsVP}
Alice Coucke, Alaa Saade, Adrien Ball, Th{\'e}odore Bluche, Alexandre Caulier,
  David Leroy, Cl{\'e}ment Doumouro, Thibault Gisselbrecht, Francesco
  Caltagirone, Thibaut Lavril, Ma{\"e}l Primet, and Joseph Dureau. 2018.
\newblock {Snips Voice Platform: an embedded Spoken Language Understanding
  system for private-by-design voice interfaces}.
\newblock \emph{arXiv preprint arXiv:1805.10190}.

\bibitem[{Desai et~al.(2020)Desai, Goh, Babu, and Aly}]{desai2020lightweight}
Shrey Desai, Geoffrey Goh, Arun Babu, and Ahmed Aly. 2020.
\newblock {Lightweight Convolutional Representations for On-Device Natural
  Language Processing}.
\newblock \emph{arXiv preprint arXiv:2002.01535}.

\bibitem[{Devlin et~al.(2019)Devlin, Chang, Lee, and
  Toutanova}]{Devlin2019BERTPO}
Jacob Devlin, Ming-Wei Chang, Kenton Lee, and Kristina Toutanova. 2019.
\newblock {BERT: Pre-training of Deep Bidirectional Transformers for Language
  Understanding}.
\newblock In \emph{Proceedings of the Conference of the North {A}merican
  Chapter of the Association for Computational Linguistics: Human Language
  Technologies (NAACL-HLT)}.

\bibitem[{Frankle and Carbin(2019)}]{frankle2019lottery}
Jonathan Frankle and Michael Carbin. 2019.
\newblock {The Lottery Ticket Hypothesis: Finding Sparse, Trainable Neural
  Networks}.
\newblock In \emph{Proceedings of the International Conference on Learning
  Representations (ICLR)}.

\bibitem[{Gao et~al.(2018)Gao, Galley, and
  Li}]{gao-etal-2018-neural-approaches}
Jianfeng Gao, Michel Galley, and Lihong Li. 2018.
\newblock {Neural Approaches to Conversational AI}.
\newblock In \emph{Proceedings of the 56th Annual Meeting of the Association
  for Computational Linguistics (NAACL): Tutorial Abstracts}.

\bibitem[{Gehring et~al.(2017)Gehring, Auli, Grangier, Yarats, and
  Dauphin}]{Gehring2017ConvolutionalST}
Jonas Gehring, Michael Auli, David Grangier, Denis Yarats, and Yann~N. Dauphin.
  2017.
\newblock {Convolutional Sequence to Sequence Learning}.
\newblock In \emph{Proceedings of the International Conference on Machine
  Learning (ICML)}.

\bibitem[{Goo et~al.(2018)Goo, Gao, Hsu, Huo, Chen, Hsu, and
  Chen}]{Goo2018SlotGatedMF}
Chih-Wen Goo, Guang Gao, Yun-Kai Hsu, Chih-Li Huo, Tsung-Chieh Chen, Keng-Wei
  Hsu, and Yun-Nung Chen. 2018.
\newblock {Slot-Gated Modeling for Joint Slot Filling and Intent Prediction}.
\newblock In \emph{Proceedings of the Conference of the North {A}merican
  Chapter of the Association for Computational Linguistics: Human Language
  Technologies (NAACL-HLT)}.

\bibitem[{Gordon et~al.(2020)Gordon, Duh, and Andrews}]{gordon2020compressing}
Mitchell~A Gordon, Kevin Duh, and Nicholas Andrews. 2020.
\newblock {Compressing BERT: Studying the Effects of Weight Pruning on Transfer
  Learning}.
\newblock \emph{arXiv preprint arXiv:2002.08307}.

\bibitem[{Guo(2018)}]{Guo2018ASO}
Yunhui Guo. 2018.
\newblock {A Survey on Methods and Theories of Quantized Neural Networks}.
\newblock \emph{arXiv preprint arXiv:1808.04752}.

\bibitem[{Gupta et~al.(2018)Gupta, Shah, Mohit, Kumar, and
  Lewis}]{gupta-etal-2018-semantic}
Sonal Gupta, Rushin Shah, Mrinal Mohit, Anuj Kumar, and Mike Lewis. 2018.
\newblock {Semantic Parsing for Task Oriented Dialog using Hierarchical
  Representations}.
\newblock In \emph{Proceedings of the Conference on Empirical Methods in
  Natural Language Processing (EMNLP)}.

\bibitem[{Han et~al.(2016)Han, Mao, and Dally}]{Han2015DeepCC}
Song Han, Huizi Mao, and William~J. Dally. 2016.
\newblock {Deep Compression: Compressing Deep Neural Network with Pruning,
  Trained Quantization and Huffman Coding}.
\newblock In \emph{Proceedings of the International Conference on Learning
  Representations (ICLR)}.

\bibitem[{Hemphill et~al.(1990)Hemphill, Godfrey, and
  Doddington}]{hemphill1990atis}
Charles~T. Hemphill, John~J. Godfrey, and George~R. Doddington. 1990.
\newblock {The ATIS Spoken Language Systems Pilot Corpus}.
\newblock In \emph{Speech and Natural Language: Proceedings of a Workshop Held
  at Hidden Valley, Pennsylvania}.

\bibitem[{Henderson(2015)}]{henderson-machine-2015}
Matthew Henderson. 2015.
\newblock {Machine Learning for Dialog State Tracking: A Review}.
\newblock In \emph{Proceedings of The First International Workshop on Machine
  Learning in Spoken Language Processing}.

\bibitem[{Hinton et~al.(2015)Hinton, Vinyals, and
  Dean}]{Hinton2015DistillingTK}
Geoffrey~E. Hinton, Oriol Vinyals, and Jeffrey Dean. 2015.
\newblock {Distilling the Knowledge in a Neural Network}.
\newblock \emph{arXiv preprint arXiv:1503.02531}.

\bibitem[{Jiao et~al.(2020)Jiao, Yin, Shang, Jiang, Chen, Li, Wang, and
  Liu}]{jiao2020tinybert}
Xiaoqi Jiao, Yichun Yin, Lifeng Shang, Xin Jiang, Xiao Chen, Linlin Li, Fang
  Wang, and Qun Liu. 2020.
\newblock {TinyBERT: Distilling BERT for Natural Language Understanding}.
\newblock \emph{arXiv preprint arXiv:1909.10351}.

\bibitem[{Kim(2014)}]{Kim2014ConvolutionalNN}
Yoon Kim. 2014.
\newblock {Convolutional Neural Networks for Sentence Classification}.
\newblock In \emph{Proceedings of the Conference on Empirical Methods in
  Natural Language Processing ({EMNLP})}.

\bibitem[{Kingma and Ba(2015)}]{Kingma2015AdamAM}
Diederik~P. Kingma and Jimmy Ba. 2015.
\newblock {Adam: A Method for Stochastic Optimization}.
\newblock \emph{Proceedings of the International Conference on Learning
  Representations (ICLR)}.

\bibitem[{Kozareva and Ravi(2018)}]{kozarevafast}
Zornitsa Kozareva and Sujith Ravi. 2018.
\newblock {Fast \& Small On-device Neural Networks for Short Text Natural
  Language Processing}.
\newblock In \emph{Proceedings of the NeurIPS Workshop on Machine Learning on
  the Phone and other Consumer Devices (MLPCD)}.

\bibitem[{LeCun et~al.(1990)LeCun, Denker, and Solla}]{lecun1990optimal}
Yann LeCun, John~S. Denker, and Sara~A. Solla. 1990.
\newblock {Optimal Brain Damage}.
\newblock In \emph{Proceedings of the Conference on Neural Information
  Processing Systems (NeurIPS)}.

\bibitem[{Li et~al.(2017)Li, Kadav, Durdanovic, Samet, and
  Graf}]{li2017pruning}
Hao Li, Asim Kadav, Igor Durdanovic, Hanan Samet, and Hans~Peter Graf. 2017.
\newblock {Pruning Filters for Efficient ConvNets}.
\newblock In \emph{Proceedings of the International Conference on Learning
  Representations (ICLR)}.

\bibitem[{Lin et~al.(2010)Lin, Gu, and Chakraborty}]{Lin2010TuningMA}
Ziheng Lin, Yan Gu, and Samarjit Chakraborty. 2010.
\newblock {Tuning Machine-Learning Algorithms for Battery-Operated Portable
  Devices}.
\newblock In \emph{Proceedings of the Asia Information Retrieval Symposium
  (AIRS)}.

\bibitem[{Liu and Lane(2016)}]{Liu2016AttentionBasedRN}
Bing Liu and Ian Lane. 2016.
\newblock {Attention-Based Recurrent Neural Network Models for Joint Intent
  Detection and Slot Filling}.
\newblock In \emph{Proceedings of the Annual Conference of the International
  Speech Communication Association (INTERSPEECH)}.

\bibitem[{McIntosh et~al.(2018)McIntosh, Hassan, and
  Hindle}]{McIntosh2018WhatCA}
Andrea~K. McIntosh, Safwat Hassan, and Abram Hindle. 2018.
\newblock {What can Android mobile app developers do about the energy
  consumption of machine learning?}
\newblock \emph{Empirical Software Engineering}.

\bibitem[{Molchanov et~al.(2016)Molchanov, Tyree, Karras, Aila, and
  Kautz}]{Molchanov2016PruningCN}
Pavlo Molchanov, Stephen Tyree, Tero Karras, Timo Aila, and Jan Kautz. 2016.
\newblock {Pruning Convolutional Neural Networks for Resource Efficient
  Inference}.
\newblock In \emph{Proceedings of the International Conference on Learning
  Representations (ICLR)}.

\bibitem[{Pennington et~al.(2014)Pennington, Socher, and
  Manning}]{pennington2014glove}
Jeffrey Pennington, Richard Socher, and Christopher Manning. 2014.
\newblock {GloVe: Global Vectors for Word Representation}.
\newblock In \emph{Proceedings of the Conference on Empirical Methods in
  Natural Language Processing (EMNLP)}.

\bibitem[{Persand et~al.(2020)Persand, Anderson, and
  Gregg}]{Persand2020CompositionOS}
Kaveena Persand, Andrew Anderson, and David Gregg. 2020.
\newblock {Composition of Saliency Metrics for Channel Pruning with a Myopic
  Oracle}.
\newblock \emph{arXiv preprint arXiv:2004.03376}.

\bibitem[{Qin et~al.(2019)Qin, Che, Li, Wen, and Liu}]{Qin2019ASF}
Libo Qin, Wanxiang Che, Yangming Li, Haoyang Wen, and Ting Liu. 2019.
\newblock {A Stack-Propagation Framework with Token-Level Intent Detection for
  Spoken Language Understanding}.
\newblock In \emph{Proceedings of the Conference on Empirical Methods in
  Natural Language Processing and the International Joint Conference on Natural
  Language Processing (EMNLP-IJCNLP)}.

\bibitem[{Rastogi et~al.(2020)Rastogi, Zang, Sunkara, Gupta, and
  Khaitan}]{rastogi2019towards}
Abhinav Rastogi, Xiaoxue Zang, Srinivas Sunkara, Raghav Gupta, and Pranav
  Khaitan. 2020.
\newblock {Towards Scalable Multi-domain Conversational Agents: The
  Schema-Guided Dialogue Dataset}.
\newblock In \emph{Proceedings of the AAAI Conference on Artificial
  Intelligence (AAAI)}.

\bibitem[{Ravi and Kozareva(2018)}]{raviselfgoverning}
Sujith Ravi and Zornitsa Kozareva. 2018.
\newblock {Self-Governing Neural Networks for On-Device Short Text
  Classification}.
\newblock In \emph{Proceedings of the Conference on Empirical Methods in
  Natural Language Processing (EMNLP)}.

\bibitem[{Sanh et~al.(2019)Sanh, Debut, Chaumond, and
  Wolf}]{Sanh2019DistilBERTAD}
Victor Sanh, Lysandre Debut, Julien Chaumond, and Thomas Wolf. 2019.
\newblock {DistilBERT, a distilled version of BERT: smaller, faster, cheaper
  and lighter}.
\newblock \emph{arXiv preprint: arXiv:1910.01108}.

\bibitem[{Srivastava et~al.(2014)Srivastava, Hinton, Krizhevsky, Sutskever, and
  Salakhutdinov}]{Srivastava2014DropoutAS}
Nitish Srivastava, Geoffrey~E. Hinton, Alex Krizhevsky, Ilya Sutskever, and
  Ruslan Salakhutdinov. 2014.
\newblock {Dropout: A Simple Way to Prevent Neural Networks from Overfitting}.
\newblock \emph{Journal of Machine Learning Research (JMLR)}.

\bibitem[{Tang et~al.(2019)Tang, Lu, Liu, Mou, Vechtomova, and
  Lin}]{tang2019distilling}
Raphael Tang, Yao Lu, Linqing Liu, Lili Mou, Olga Vechtomova, and Jimmy Lin.
  2019.
\newblock {Distilling Task-Specific Knowledge from BERT into Simple Neural
  Networks}.
\newblock \emph{arXiv preprint arXiv:1903.12136}.

\bibitem[{Tjong and Sang(2000)}]{sang-2000-bio}
Erik~F. Tjong and Kim Sang. 2000.
\newblock Transforming a chunker to a parser.
\newblock In \emph{Proceedings of the Meeting of Computational Linguistics in
  the Netherlands (CLIN)}.

\bibitem[{Vaswani et~al.(2017)Vaswani, Shazeer, Parmar, Uszkoreit, Jones,
  Gomez, Kaiser, and Polosukhin}]{Vaswani2017AttentionIA}
Ashish Vaswani, Noam Shazeer, Niki Parmar, Jakob Uszkoreit, Llion Jones,
  Aidan~N. Gomez, Lukasz Kaiser, and Illia Polosukhin. 2017.
\newblock {Attention is All you Need}.
\newblock In \emph{Proceedings of the Conference on Neural Information
  Processing Systems (NeurIPS)}.

\bibitem[{Wang et~al.(2018)Wang, Shen, and Jin}]{Wang2018ABB}
Yu~Wang, Yilin Shen, and Hongxia Jin. 2018.
\newblock {A Bi-Model Based RNN Semantic Frame Parsing Model for Intent
  Detection and Slot Filling}.
\newblock In \emph{Proceedings of the Conference of the North American Chapter
  of the Association for Computational Linguistics: Human Language Technologies
  (NAACL-HLT)}.

\bibitem[{Wr{\'o}bel et~al.(2018)Wr{\'o}bel, Pietroń, Wielgosz, Karwatowski,
  and Wiatr}]{Wrbel2018ConvolutionalNN}
Krzysztof Wr{\'o}bel, Marcin Pietroń, Maciej Wielgosz, Michal Karwatowski, and
  Kazimierz Wiatr. 2018.
\newblock {Convolutional neural network compression for natural language
  processing}.
\newblock \emph{arXiv preprint arXiv:1805.10796}.

\bibitem[{Yan et~al.(2017)Yan, Duan, Chen, Zhou, Zhou, and
  Li}]{Yan2017BuildingTD}
Zhao Yan, Nan Duan, Peng Chen, Ming Zhou, Jianshe Zhou, and Zhoujun Li. 2017.
\newblock {Building Task-Oriented Dialogue Systems for Online Shopping}.
\newblock In \emph{Proceedings of the AAAI Conference on Artificial
  Intelligence (AAAI)}.

\bibitem[{Yoon et~al.(2018)Yoon, Robinson, Christian, Qiu, and
  Tourassi}]{yoon2018}
Hong-Jun Yoon, Sarah Robinson, J.~Blair Christian, John~X. Qiu, and Georgia~D.
  Tourassi. 2018.
\newblock {Filter pruning of Convolutional Neural Networks for text
  classification: A case study of cancer pathology report comprehension}.
\newblock In \emph{Proceedings of the IEEE EMBS International Conference on
  Biomedical \& Health Informatics (BHI)}.

\bibitem[{Yu et~al.(2016)Yu, Xu, Black, and Rudnicky}]{yu-etal-2016-strategy}
Zhou Yu, Ziyu Xu, Alan~W Black, and Alexander Rudnicky. 2016.
\newblock {Strategy and Policy Learning for Non-Task-Oriented Conversational
  Systems}.
\newblock In \emph{Proceedings of the Annual Meeting of the Special Interest
  Group on Discourse and Dialogue (SIGDIAL)}.

\bibitem[{Zafrir et~al.(2019)Zafrir, Boudoukh, Izsak, and
  Wasserblat}]{zafrir2019q8bert}
Ofir Zafrir, Guy Boudoukh, Peter Izsak, and Moshe Wasserblat. 2019.
\newblock {Q8BERT: Quantized 8Bit BERT}.
\newblock \emph{arXiv preprint arXiv:1910.06188}.

\bibitem[{Zhang et~al.(2019)Zhang, Li, Du, Fan, and Yu}]{zhang2019joint}
Chenwei Zhang, Yaliang Li, Nan Du, Wei Fan, and Philip Yu. 2019.
\newblock {Joint Slot Filling and Intent Detection via Capsule Neural
  Networks}.
\newblock In \emph{Proceedings of the Annual Meeting of the Association for
  Computational Linguistics (ACL)}.

\bibitem[{Zhang et~al.(2020)Zhang, Takanobu, Huang, and Zhu}]{zhang2020recent}
Zheng Zhang, Ryuichi Takanobu, Minlie Huang, and Xiaoyan Zhu. 2020.
\newblock {Recent Advances and Challenges in Task-oriented Dialog System}.
\newblock \emph{arXiv preprint arXiv:2003.07490}.

\end{thebibliography}
\bibliographystyle{acl_natbib}

% \appendix

\end{document}